\documentstyle[epsfig]{article}
\def\ArtWork#1{\noindent\hfill\epsfbox{#1}\hfill}
\newcommand{\btb}{\begin{tabular}}
\newcommand{\etb}{\end{tabular}}
\newcommand{\bd}{\begin{document}}
\newcommand{\ed}{\end{document}}
\newcommand{\bc}{\begin{center}}
\newcommand{\ec}{\end{center}}
\newcommand{\vs}{\vspace}
\newcommand{\hs}{\hspace}
\newcommand{\bq}{\begin{quote}}
\newcommand{\eq}{\end{quote}}
\newcommand{\lb}{\linebreak}
\newcommand{\pb}{\pagebreak}
\newcommand{\mb}{\makebox}
\newcommand{\bit}{\begin{itemize}}
\newcommand{\eit}{\end{itemize}}
\newcommand{\ben}{\begin{enumerate}}
\newcommand{\een}{\end{enumerate}}
\newcommand{\sbs}{\subset}
\newcommand{\lt}{\left}
\newcommand{\rt}{\right}
\newcommand{\tm}{\times}
\newcommand{\hf}{\hspace*{\fill}}
\newcommand{\vf}{\vspace*{\fill}}
\newcommand{\beq}{\begin{equation}}
\newcommand{\eeq}{\end{equation}}
\newcommand{\ba}{\begin{array}}
\newcommand{\ea}{\end{array}}
\newcommand{\beqa}{\begin{eqnarray}}
\newcommand{\eeqa}{\end{eqnarray}}
\newcommand{\beqas}{\begin{eqnarray*}}
\newcommand{\eeqas}{\end{eqnarray*}}
\newcommand{\pad}{\partial}
\newcommand{\bm}[1]{\hbox{{\mb{\boldmath ${#1}$}}}}
\newcommand{\fb}[1]{\lt[{#1}\rt]}
\newcommand{\ot}{\otimes}
\newcommand{\nn}{\nonumber}
\newcommand{\<}{\langle}
\renewcommand{\>}{\rangle}
\newcommand{\R}{\mb{$I\!\!R$}}
\newcommand{\C}{{\cal C}}
\newcommand{\M}{{\cal M}}
\newcommand{\N}{{\cal N}}
\newcommand{\A}{{\cal A}}
\renewcommand{\P}{{\cal P}}
\newcommand{\es}{\emptyset}
\newcommand{\ci}{\subseteq}
\newcommand{\cs}{\supseteq}
\renewcommand{\u}{\cup}
\renewcommand{\i}{\cap}
\newcommand{\bu}{\bigcup}
\newcommand{\bi}{\bigcap}
\newcommand{\la}{\leftarrow}
\newcommand{\ra}{\rightarrow}
\newcommand{\Ra}{\Rightarrow}
\newcommand{\Lra}{\Leftrightarrow}
\newcommand{\lgra}{\longrightarrow}
\newcommand{\Lgra}{\Longrightarrow}
\newcommand{\lglra}{\longleftrightarrow}
\newcommand{\Lglra}{\Longleftrightarrow}
\renewcommand{\a}{\alpha}
\renewcommand{\b}{\beta}
\newcommand{\g}{\gamma}
\newcommand{\G}{\Gamma}
\renewcommand{\d}{\delta}
\renewcommand{\th}{\theta}
\newcommand{\e}{\varepsilon}
\newcommand{\eps}{\epsilon}
\newcommand{\h}{\eta}
\renewcommand{\l}{\lambda}
\newcommand{\m}{\mu}
\newcommand{\n}{\nu}
\newcommand{\p}{\pi}
\newcommand{\s}{\sigma}
\newcommand{\Si}{\Sigma}
\newcommand{\ta}{\tau}
\newcommand{\ph}{\phi}
\newcommand{\Ph}{\Phi}
\renewcommand{\c}{\chi}
\newcommand{\om}{\omega}
\newcommand{\Om}{\Omega}
\newcommand{\tri}{\triangle}
\newcommand{\rec}[1]{\frac{1}{#1}}
\newcommand{\f}{\frac}
\newcommand{\sm}[2]{\sum_{#1}^{#2}}
\newcommand{\ld}{\ldots}
\newcommand{\ov}{\overline}
\newcommand{\un}{\underline}
\newcommand{\iy}{\infty}
\renewcommand{\S}{{\cal S}}
\newcommand{\qed}{\hf\rule{3mm}{3mm}}
\newcommand{\wt}{\widetilde}
\newcommand{\ds}{\displaystyle}
\newcommand{\bdm}{\begin{displaymath}}
\newcommand{\edm}{\end{displaymath}}
\newcommand{\et}[2]{#1_{1},#1_{2},\ld  ,#1_{#2}}
\newcommand{\alter}[2]{\lt\{ \ba {ll}#1 \\ #2 \ea \rt.}
\newcommand{\alt}[4]{\lt\{ \ba{ll}#1 & \mb{if \,\,}#2 \\ #3 & \mb{if
               \,\,}#4 \ea \rt.}
\newcommand{\altn}[4]{\lt\{ \ba{rl}#1 & \mb{if \,\,}#2 \\ #3 & \mb{if
               \,\,}#4 \ea \rt.}
\newcommand{\alto}[6]{ \lt\{ \ba{ll}#1 & \mb{if \,\,}#2 \\ #3 & \mb{if
               \,\,} #4 \\ #5 & \mb{if \,\,}#6 \ea \rt.}
\newcommand{\altero}[5]{\mb{$\lt\{ \ba {ll}#1 & \mb{if \,\,}#2 \\ #3 &
               \mb{if \,\,} #4 \\ #5 & \mb{otherwise} \ea \rt.$}}

\newcommand{\blr}{\begin{list}{~\roman{romc})} {\usecounter{romc}
                \setlength{\topsep}{0pt} \setlength{\itemsep}{0pt}}}
\newcommand{\elr}{\end{list}}
\newcounter{romc}
\newcommand{\bx}{{\bf x}}
\newcommand{\bw}{{\bf w}}

\renewcommand{\textwidth}{450pt}
\renewcommand{\textheight}{700pt}
\renewcommand{\baselinestretch}{1.0}

\begin{document}
\setlength{\parskip}{0.5pc}
\setlength{\parindent}{0.0cm}
\addtolength{\oddsidemargin}{-2cm}
\addtolength{\topmargin}{-2.5cm}

\title{Land cover classification using fuzzy rules and aggregation of
contextual information through evidence theory\footnote{IEEE Transactions on Geoscience and Remote Sensing, Vol. 44, No. 6, pp. 1633-1642, June 2006.}}

\author{
Arijit Laha\thanks{Arijit Laha is
with Institute for Development and Research in Banking Technology,
Castle Hills, Hyderabad 500 057, India, E-mail: arijit.laha@ieee.org},
Nikhil R. Pal, and J. Das
\thanks{Nikhil R. Pal and J. Das are with Electronics and Communication
Science Unit, Indian Statistical Institute, 203 B. T. Road
Calcutta 700 108, India, E-mail: \{nikhil, jdas\}@isical.ac.in} }
\date{}
\maketitle

\begin{abstract}Land cover classification using multispectral
satellite image is a very challenging task with numerous practical
applications. We propose a multi-stage classifier that involves
fuzzy rule extraction from the training data and then generation
of a possibilistic label vector for each pixel using the fuzzy rule
base. To exploit the spatial correlation of land cover types we
propose four different information aggregation methods which use
the possibilistic class label of a pixel and those of its eight
spatial neighbors for making the final classification decision.
Three of the aggregation methods use Dempster-Shafer theory of
evidence while the remaining one is modeled after the fuzzy k-NN
rule. The proposed methods are tested with two benchmark seven
channel satellite images and the results are found to be quite
satisfactory. They are also compared with a Markov random field
(MRF) model-based contextual classification method and found to
perform consistently better.
\end{abstract}


\section{Introduction}

Land cover classification in remotely sensed images is considered
to be a cost effective and reliable method for generating
up-to-date land cover information \cite{foody}. Usually such
images are captured by multispectral scanners (such as Landsat
TM), that acquire data at several distinct spectral bands
producing multispectral images.  Automated analysis of such data
calls for sophisticated techniques for data fusion and pattern
recognition. Most widely used techniques include statistical
modeling involving discriminant analysis and maximum likelihood
classification and artificial neural networks-based approaches
\cite{paola}, \cite{pinz}, \cite{ji}, \cite{foody2}. However,
these classifiers usually classify a sample to the class for which
maximum support is obtained, no matter how small this amount of
support may be or there may be another class for which the support
is very close to the maximum. This particular feature is often
criticized \cite{foody} in context of their applicability in land
cover classification. ``Soft" classifiers can be useful for such
problems as they can produce a measure of confidence in support of
the decision as well as indicate measures of confidence in support
of alternative decisions, which can be used for further processing
 using auxiliary information. This can result in a
more robust and accurate system. In developing soft classifiers
for land cover analysis two approaches have gained popularity.
These are  based on (1)fuzzy set theory \cite{bez2} and (2)
Dempster and Shafer's evidence theory \cite{shafer}. There is, of
course, the probabilistic approach, that we do not pursue further.

Different fuzzy methodologies for land cover classification in
multispectral satellite images have been investigated by various
researchers. For example, \cite{foody} uses fuzzy $c$-means
algorithm, while Kumar et al. \cite{kumar2} applied a fuzzy integral
method. Fuzzy rulebased systems have been used for classification by
many researchers \cite{bez2,ludmila} for diverse fields of
application. Fuzzy rules are attractive because they are
interpretable and can provide an analyst a deeper insight into the
problem. Use of fuzzy rulebased systems for land cover analysis is a
relatively new approach. Recently B\'{a}rdossy and Samaniego
\cite{bard} have proposed a scheme for developing a fuzzy rulebased
classifier for analysis of multispectral images, where the randomly
generated initial rules are fine-tuned by simulated annealing. In
\cite{kulkarni} Kulkarni and McCaslin used fuzzy neural networks for
rule extraction.

The other approach to design soft classifiers use the evidence
theory developed by Dempster and Shafer \cite{shafer},\cite{deno}, \cite{zouhal}.
 As observed by
Lee et al. in \cite{lee}, for multispectral image analysis there
may be a great incentive for applying Dempster-Shafer theory of
evidence. Since the theory of evidence allows one to combine
evidences obtained from diverse sources of information in support
of a hypothesis, it seems a natural candidate for analyzing
multispectral images for land cover classification\cite{srini}, \cite{peddle}, \cite{kim}.
In all
these works the approach is to treat each channel image as a
separate source of information. Each image is analyzed to
associate each pixel with some degree of belief pertaining to its
belonging to each member of a set of hypotheses known as the {\em
frame of discernment}. Usually some probabilistic techniques are
employed to assign the degree of belief. In the next stage these
belief values from all images for a pixel are combined using Dempster's
rule \cite{shafer} to calculate the total support for each
hypothesis. In a recent paper \cite{jouan} Jouan and Allard used
evidence theory for combining information from multiple sources
for land use mapping.

In a satellite image, usually the landcover classes form spatial
clusters, i.e., a pixel belonging to a particular class is more
likely to have neighboring pixels from the same class rather than
from other classes. Thus, inclusion of contextual information from
the neighboring pixels is likely to increase classification
accuracy. An overview of common contextual pattern recognition
methods can be found in \cite{haralick}. In this paper we
propose several schemes for classifier design that uses both fuzzy sets
theory and Dempster-Shafer evidence theory. These are two stage
schemes. In the first stage a fuzzy rulebased classifier is
developed using a small   training set.
This classifier is noncontextual and for each pixel it generates a
possibilistic label vector. In the next stage we aggregate the
responses of the fuzzy rules over a $3\times 3$ {\em spatial}
neighborhood of a pixel to make the classification decision about
that pixel. Thus, the decision making process takes into
consideration the information available from the spatial
neighborhood of the pixel.
Here we propose four methods for contextual decision making. In
the experimental results we compare the performance of the
contextual classifiers with the fuzzy rulebased noncontextual
classifiers designed in the first stage. We also provide a
comparison of the performance of the proposed classification
schemes with a recent Markov random field (MRF) model-based
contextual classification scheme proposed by Sarkar et al.
\cite{sarkar}.

\section{Designing the Fuzzy Rule base}

We use a multi-stage scheme for designing fuzzy rule based
classifiers. The set of gray values corresponding to a pixel in
the channel images is used as the feature vector for that pixel.
In the first stage a set of labeled prototypes representing the
distribution of the training data is generated using a
Self-organizing Feature Map (SOFM) \cite{k1} based algorithm
developed in \cite{laha2}. The algorithm dynamically decides the
number of prototypes based on the training data. Then each of
these prototypes is converted to a fuzzy rule.

Next the fuzzy rules are tuned by modifying the peaks as well as
the spreads of the fuzzy sets associated with the rules. The tuned
rules can readily be used to classify unknown samples based on the
firing strengths of the rules. Typically, a test sample is
classified to the class of the rule generating the highest firing
strength. For the sake of completeness we first give a brief
description of SOFM.

\subsection{Kohonen's SOFM algorithm}

SOFM is formed of neurons placed on a regular (usually) 1D or 2D
grid. Thus each neuron is identified with a index corresponding to
its position in the grid (the viewing plane). Each neuron $i$ is
represented by a weight vector ${\bf w}_i \in \Re^p$ where $p$ is
the dimensionality of the input space. In $t$-th training step, a
data point ${\bf x}\in \Re^p$ is presented to the network. The
winner node with index $r$ is selected as $r = \underbrace{arg\
min}_{i} \{\|{\bf x} - {\bf w}_{i,t-1}\|\}$. ${\bf w}_{r,t-1}$ and
the other weight vectors associated with cells in the spatial
neighborhood $N_{t}(r)$ of $r$ are updated using the rule: $$ {\bf
w}_{i,t} = {\bf w}_{i,t-1} + \a(t)h_{ri}(t)({\bf x} - {\bf
w}_{i,t-1}),$$ where $\a(t)$ is the learning rate and $h_{ri}(t)$
is the neighborhood kernel (usually Gaussian). The learning rate
and the radius of the neighborhood kernel decrease monotonically
with time. During the iterative training the SOFM behaves like a
flexible net that folds onto the ``cloud" formed by the input
data. A trained SOFM exhibits remarkable and useful properties of
{\em topology preservation} and {\em density matching} and often
used for visualization of metric-topological relationships and
distributional density properties of training data $X = \{{\bf
x}_{1},...,{\bf x}_{N}\}$ in $\Re^{p}$ through their mapping onto
the viewing plane. From clustering viewpoint, SOFM has the
advantage of avoiding under-utilization of the prototypes.

\subsection{Generation of Prototypes}

First we train a one dimensional SOFM with $c$ nodes, where $c$ is
the number of classes.We do so because the smallest number of rules
that may be required is equal to the number of classes. At the end
of the training the weight vector distribution of the SOFM reflects
the distribution of the input data. Then the training data is
divided into a $c$ partition according to their closeness to $c$
weight vectors, $V^{0} = \{ {\bf v}_1^0,{\bf v}_2^0, ..., {\bf
v}_c^0\} \subset \Re^{p}$. Each of the prototypes is labeled based on
the class information of corresponding partition using {\em majority
voting}. However, such a set of prototypes may not classify the data
satisfactorily because the class information is not used in the
training resulting in the possibility that a prototype may represent
data from more than one class significantly.

We use the prototype refinement scheme described in \cite{laha2}.
The basic idea behind this refinement algorithm is that a useful
prototype should satisfy two criteria: (i) it should represent an
adequate number of points, (ii) only one of the classes should be
strongly represented by it.

The prototype refinement scheme applies four operations, {\em
deletion}, {\em modification}, {\em merging} and {\em splitting} on
the set of prototypes while trying to fulfill the above conditions.
 In the training data $X = \{{\bf
x}_{1},...,{\bf x}_{N}\}$ let there be $N_{j}$ points from class
$j$. The refinement stage uses just {\em two} parameters $K_{1}$ and
$K_{2}$ to dynamically generate $c+1$ retention thresholds known as
a global retention threshold $\alpha$ and a set of class-wise
retention threshold $\beta_{k}$ (one for each class), to evaluate
the performance of each prototype. $\alpha$ and $\beta_{k}$ are
computed dynamically (not fixed) for the $t$-th iteration using the
following formulae:
$$
\alpha ^t  = \left( {K_1 \left| {V^{t - 1} } \right|} \right)^{ -
1} \hbox{  and  }\beta_k^t = \left( {K_{2} \left| V_k^{t-1}
\right|}\right)^{ - 1},$$ where $ V_k^{t-1} = \{ {\bf v}_{i} \mid
{\bf v}_{i} \in V^{t-1}, C_{i} = k\}$.


Here $ V^{t-1}$ is the set of prototypes obtained after $(t-1)$
iterations of the algorithm. To consider a prototype useful it must
represent more than $\alpha^{t}N$ training points. Further, a
prototype must represent more than $\beta_k^tN_k$ points from class
$k$ to be considered a (potential) prototype for class $k$.

Finally, the set of prototypes is again refined by SOFM algorithm
with winner-only update strategy. After a few iterations this
algorithm produces a set of adequate number of prototypes that
represents the training data much better than the initial one. For
details the readers are referred to \cite{laha2}. The final set of
prototypes $V^{final}=\{{\bf v}_1^{final}, \cdots, {\bf v}_{\hat
c}^{final} \}$, where ${\hat c}\geq c$, is used to generate the set
of initial fuzzy rules.

\subsection{Designing fuzzy rulebase}
A prototype ${\bf v}_i$ represents a cluster of points for class
$k$. This cluster can be described by a fuzzy rule of the form:
$R_{i}$: If ${\bf x}$ is CLOSE TO ${\bf v}_i$  then class is $k$.
  This rule can be further  translated into  :
\begin{description}
    \item[$R_{i}$:] If $x_1$ is CLOSE TO $v_{i1}$ AND $\cdots$ AND $x_p$ is
CLOSE TO $v_{ip}$ then class is $k$.
\end{description}
Note that, this is just one possible interpretation of ``${\bf x}$
is CLOSE TO ${\bf v}_i$". The first form requires a multidimensional
membership function while the second form requires several one
dimensional membership functions and a conjunction operator. In
general, the two forms will not produce the same output. Depending
on the choice of the membership function and the conjunction
operator, the forms may lead to the same output.  But, since the
representation by the atomic clauses is a plausible realization of
the multidimensional form, the performance of the systems built
using either form is not going to be much different.

The fuzzy set   CLOSE TO $v_{ij}$  can be modeled by triangular,
trapezoidal or Gaussian membership function. In this
investigation, we use the Gaussian membership function,
$$\mu_{ij}  ( x_j; v_{ij},\sigma_{ij})=\exp({-{(x_j -
v_{ij})}^2/{\sigma_{ij}}^2}).$$ Given a   data point ${\bf x}$ with
unknown class, we first find the firing strength of each rule. Let
$\alpha_i({\bf x})$ denote the firing strength of the $i^{th}$ rule
on a data point ${\bf x}$. We assign the point ${\bf x}$ to class
$k$ if $\alpha_r = \max_i(\alpha_i({\bf x}))$ and the $r^{th}$ rule
represents class $k$.

The performance of the classifier depends crucially on the adequacy
of the number of rules used and proper choice of the membership
functions. In our case each fuzzy set is characterized by two
parameters $v_{ij}$ and $\sigma_{ij}$. Let the initial set of fuzzy
rules be $R^0=\{R_i^0\mid i=1,2,\ldots, {\hat c}\}$. The parameters
$v_{ij}^0$ and $\sigma_{ij}^0$ for fuzzy sets in the antecedent part
of a rule $R_i^0\in R^0$ are obtained from the prototype ${\bf
v}_i^{final}\in V^{final}$ as follows:
$$v_{ij}^0=v_{ij}^{final}\hbox{ and }\sigma_{ij}^0
=k_{w}(\sqrt(\sum_{{\bf x}_{k} \in
X_{i}}(x_{kj}-v_{ij}^{final})^2)/|X_{i}|,$$ where $X_{i}$ is the set
of training data closest to ${\bf v}_{i}^{final}$ and $k_{w} > 0$ is
a constant parameter that controls the initial width of the
membership function. If $k_{w}$ is small, then specificity of the
fuzzy sets defining the rules will be high and hence each rule will
model a small area in the input space. On the other hand, a high
$k_{w}$ will make each rule cover a bigger area.  Since the spreads
are tuned, in principle $k_{w}$ should not have much impact on the
final performance, but in practice   the  value of $k_{w}$ may  have
a significant impact on the classification performance for
complicated data sets because of the local minimum problem of
gradient descent techniques. In the current work the values of
$k_{w}$s are found experimentally. One can use a validation set for
this.

The initial rulebase $R^0$ thus obtained can be further fine tuned
to achieve better performance. But the exact tuning algorithm
depends on the conjunction operator (implementing AND operation for
the antecedent part) used for computation of the firing strengths.
The firing strength can be calculated using any T-norm \cite{bez2}.
Use of different T-norms results in different classifiers. The {\bf
product} and the {\bf minimum} are the most popular choices of
T-norms.

Though it is much easier to formulate a calculus based tuning
algorithm if product is used, its use is conceptually somewhat
unattractive. To illustrate the point let us consider a rule
having two atomic clauses in its antecedent. If the two clauses
have truth values $a$ and $b$, then intuitively the antecedent is
satisfied at least to the extent of $\min(a,b)$. However, if
product is used as the conjunction operator, we always have $ab
\leq \min(a,b)$. Thus we always under-determine the importance of
the rule. This does not cause any problem for non-classifier fuzzy
systems because the defuzzification operator usually performs some
kind of normalization with respect to the firing strength. But in
classifier type applications a decision may appear to be made
with a very low confidence, when actually it is  not the case. In
certain cases, such a situation may lead to overemphasis on total
ignorance under evidence theory framework. Thus to avoid the use
of the product and at the same time to be able to apply calculus
to derive update rules we use a soft-min operator.

The {\bf soft-match} of $n$ positive number
$x_1,x_2,...,x_n$ is defined by
$$SM(x_1,x_2,...,x_n,q) = \left\{\frac{(x_1^q+x_2^q+...+x_n^q)}{n}\right\}^{1/q}.$$
where $q$ is any real number. $SM$ is known as an aggregation
operator with upper bound of value 1 when $x_i \in [0,1] \forall
i$. This operator is used by different authors \cite{dyck,chinmoy}
for different purposes. It is easy to see that as $q\to -\infty$
the soft-match operator behaves like ``min" operator. Thus we
define the softmin operator as the soft match operator with a
sufficiently negative value of the parameter $q$. The firing
strength of the r-th rule computed using softmin is
$$\alpha_r({\bf x}) = \left\{\frac{\sum_{j=1}^{j=p}(\mu_{rj}(x_j; v_{rj}, \sigma_{rj}))^q}{p}\right\}^{1/q}.$$
At the value $q=-10$ its behavior resembles very closely to the min
operator, so we use $q = -10.0$. A very low value of $q$ may lead to
numerical instability in the computation.

\subsection{Tuning of the Rule Base}
Let ${\bf x} \in X$ be from class $c$ and ${ R}_{c}$ be the rule
from class $c$ giving the maximum firing strength $\a_{c}$ for
${\bf x}$. Also let ${R}_{\neg c}$ be the rule from the incorrect
classes having the highest firing strength $ \a_{\neg c}$ for
${\bf x}$. We use the error function $E$, \beq E = \sum_{{\bf x}
\in X}(1-\a_{c}+\a_{\neg c})^{2}. \eeq

This kind of error function has been used in \cite{chiu}. We
minimize $E$ with respect to $v_{cj}$, $v_{\neg cj}$ and
$\sigma_{cj}$, $\sigma_{\neg cj}$ of the two rules $R_{c}$ and
$R_{\neg c}$ using gradient decent. Here the index $j$ corresponds
to clause number in the corresponding rule. The tuning process is
repeated until the rate of decrement in $E$ becomes negligible
resulting in the rule base $R^{final}$.

Since a Gaussian membership function is extended to infinity, for
any data point all rules will be fired to some extent. In our
implementation, if the firing strength is less than a threshold,
$\epsilon$ $(\approx 0.01)$, then the rule is not assumed to be
fired. The threshold is set considering approximate $2\sigma$
limit of the Gaussian membership functions. Thus, under this
situation, the rulebase extracted by the system may not be
complete with respect to the training data. This can happen even
when we use membership functions with triangular or trapezoidal
shapes. This is not a limitation but a distinct advantage,
although for the data sets we used, we did not encounter such a
situation. If  no rule is fired by a  data point, then that point
can be thought of as an outlier. If such a situation occurs for
some test data, then that will indicate an observation not close
enough to the training data and consequently no conclusion should
be made about such test points.

\section{Decision making with aggregation of spatial contextual information}
In this section we describe four decision making schemes. The
first one involves a simple aggregation of the fuzzy label vectors
of the neighboring pixels. The scheme is modeled after the well
known fuzzy k-nearest neighbor algorithm proposed by Keller et al.
\cite{keller}. The other three methods use Dempster-Shafer theory
of evidence. For the sake of completeness we give a brief
introduction of Dempster-Shafer theory of evidence before we
describe the decision making methods in detail.

\subsection{Dempster-Shafer theory of evidence}
Let $\Theta$ be the universal set and $P(\Theta)$ be its power set.
A function $m : P(\Theta) \rightarrow [0,1]$ is called a Basic
Probability Assignment (BPA) whenever $m(\emptyset) = 0$ and
$\sum_{A\subseteq \Theta} m(A) =1$. Here $m(A)$ is interpreted as
the degree of evidence supporting the claim that the ``truth" is in
$A$ and in absence of further evidence no more specific statement
can be made. Every set $A\in P(\Theta)$ for which $m(A)
> 0$ is called a {\em focal element} of $m$. Evidence obtained in
the same context from two distinct sources and expressed by two BPAs
$m^1$ and $m^2$ on the same power set $P(\Theta)$ can be combined by
Dempster's rule of combination to obtain a joint BPA $m^{1,2}$ as:
\beq m^{1,2}(A) = \left\{\begin{array}{ll}
            \frac{\sum_{B\cap C=A}m^1(B)m^2(C)}{1-K} &\mbox{if $A\neq \emptyset$}\\
            0 &\mbox{if $A=\emptyset$}
            \end{array}
        \right.\eeq
Here $K = \sum_{B\cap C = \emptyset}m^1(B)m^2(C).$

Equation (2) is often expressed with the notation $m^{1,2}=m^1\oplus
m^2$. The rule is commutative and associative. Evidence from any
number (say $k$) of distinct sources can be combined by repetitive
application of the rule as $m = m^1\oplus m^2\oplus\cdots\oplus m^k
=\oplus_{i=1}^km^i.$

To select the optimal decision from the evidence embodied in a BPA,
typically we construct a probability function $P^\Theta$ on $\Theta$.
This is done through a transformation known as {\em pignistic
transformation} \cite{smets}. The pignistic probability for $\theta
\in \Theta$ can be expressed in terms of BPAs as follows: \beq
P^\Theta(\theta) = \sum_{A\subseteq \Theta, \theta\in
A}\frac{m(A)}{\mid A\mid}\eeq

Optimal decision can now be chosen in favor of $\theta_0$ with the
highest pignistic probability.

\subsection{Aggregation of context information for decision making}
Let there be $c$ classes, $\cal{C}$=$\{C_1,C_2,\cdots C_c\}$. For
each pixel $P$ the fuzzy rule base generates a fuzzy label vector
${\bm \a} \in \Re^c$ such that the value of the $k$-th component
of ${\bm \a}$, $\a_k$ ($0\leq\a_k\leq1$) represents the confidence
of the rule base supporting the fact that the pixel $P$ belongs to
class $C_k$. Strictly speaking, $\bm{\a} \in \Re^c$ is a
possibilistic label vector \cite{bez2}. The value of $\a_k$ is
computed as follows:\hf\lb Let $r$ be the number of rules in the
fuzzy rulebase. Since $c\leq r$, there could be multiple rules
corresponding to a class. Then $\a_k$ is the highest firing
strength produced by the rules corresponding to the class $C_k$
for pixel $P$. We treat this value as the confidence measure of
the rulebase pertaining to the membership of pixel $P$ to the
class $C_k$. However, if $\a_k$ is less than a threshold, say
0.01, it is set to 0.

In our decision making schemes we consider a pixel together with
the pixels within its $3\times 3$ spatial neighborhood. We
identify the pixel $p$ under consideration as $p^0$ and its eight
neighbors as $p^1, p^2,\cdots p^8$. The corresponding
possibilistic label vectors assigned to these nine pixels are
denoted as $\bm{\a}^0,\bm{\a}^1, \cdots \bm{\a}^8$.

\subsubsection{Method 1: Aggregation of possibilistic labels}
This is a simple aggregation scheme modeled after the fuzzy k-NN
method of Keller et al. \cite{keller}. Given a set of nine label
vectors $\{\bm{\a}^0,\bm{\a}^1, \cdots \bm{\a}^8\}$ an aggregated
possibilistic label vector $\bm{\a}^A$ is computed as follows:
\beq \bm{\a}^A = \frac{\sum_{i=0}^8\bm{\a}^i}{9}.\eeq The pixel
$p^0$ is assigned to class $C_k$, such that
$\bm{\a}^A_k\geq\bm{\a}^A_i\hs*{1mm}\forall i=1,2,\cdots c.$ Note
that, though the label vector $\bm{\a}^0$ corresponds to the pixel
to be classified, no special emphasis (importance) is given to it
in this aggregation scheme.

\subsubsection{Method 2: Aggregation with Bayesian belief modeling}
This aggregation method as well as the next two use the evidence
theoretic framework of Dempster-Shafer. Within this framework the
set of classes, $\cal{C}$ is identified as the {\em frame of
discernment}. A body of evidence is represented by a BPA $m$ over
subsets of $\cal{C}$.
The value $m(A)$ denotes the belief in support of the proposition
that {\em the true class label of the pixel of interest is in $A
\subset \cal{C}$}. In context of our problem we shall (1) identify
distinct bodies of evidence (BOE), (2) formulate a realistic scheme
of assigning the BPAs to the relevant subsets of $\cal{C}$, (3)
combine evidences provided by all BOEs, (4) compute the pignistic
probability for each class from the combined evidence, and (5) make
a decision using the maximum pignistic probability principle.

In this scheme we identify eight BOEs for eight neighbor pixels
with corresponding BPAs denoted as $m^1,m^2,\cdots ,m^8$. We
consider the Bayesian belief structure, i.e., each focal element
has only one element. Assigning BPA to a subset essentially
involves committing some portion of belief in favor of the
proposition represented by the subset. So the scheme followed for
assigning BPAs must reflect some realistic assessment of the
information available in favor of the proposition. We define $m^i$
as follows:
 \beq m^i(\{C_k\}) = \frac{(\a_k^i + \a_k^0)}{S},
k=1,2,...,c\eeq where $S = \sum_{k=1}^{k=c}(\a_k^i + \a_k^0)$ is a
normalizing factor. Thus, each BPA contains $c$ focal elements,
one corresponding to each class and the assigned value
$m^i(\{C_k\})$ is influenced by the magnitudes of firing strengths
produced by the rule base in support of class $C_k$ for the pixel
of interest $p^0$ and its $i$-th neighbor $p^i$. Clearly the label
vector $\bm{\a}^0$ influences all eight BPAs. Hence it is expected
that in the final decision making, the influence of $\bm{\a}^0$
will be higher than any neighbor. Such an assignment is motivated
by the fact, the spatial neighbors usually are highly correlated,
i.e., pixel $p^0$ and its immediate neighbors are expected from
the same class.

Thus for eight neighboring pixels we obtain eight separate BPAs.
These BPAs are combined to get the global BPA
 applying the Dempster's rule repeatedly.  It can be
easily seen that the global BPA $m^G$ is also Bayesian and can be
computed as \beq m^G(\{C_k\}) =
\frac{\prod_{i=1}^8m^i(\{C_k\})}{\sum_{l=1}^c\prod_{i=1}^8m^i(\{C_l\})}.
\eeq

In this case the pignistic probability $P^{\cal{C}}(C_k)$ is the
same as $m^G(\{C_k\})$. So the pixel $p^0$ is assigned to class
$C_k$ such that $m^G(\{C_k\})\geq m^G(\{C_l\})\hs*{1mm}\forall
C_l\in \cal{C}.$

\subsubsection{Method 3: Aggregation with non-Bayesian belief}
Here also the BOEs are identified in the same way as in the
previous method. However, we allow the assignment of belief to
subsets of $\cal{C}$ having two elements i.e., the subsets
$\{C_l,C_m : l<m \hbox{ and } l,m = 1,2, \cdots,c\}$. We define
$m^i$ as follows:

\beq m^i(\{C_k\}) = \frac{(\a_k^i + \a_k^0)}{S}, k=1,2,...,c\eeq

\beq m^i(\{C_l,C_m : l<m\}) = \frac{(\a_l^i + \a_m^0) + (\a_m^i +
\a_l^0)}{2S}, \eeq
 $l,m=1,2,...,c,$
where \[S = \sum_{k=1}^{k=c}(\a_k^i + \a_k^0) +
    \sum_{l=1}^{l=c-1}\sum_{m=l+1}^{m=c}[(\a_l^i + \a_m^0) + (\a_m^i + \a_l^0)].\]

In Method 2 we exploited only the correlation of spatial
neighbors. However, for satellite images when a pixel falls at the
boundary of some land cover type, it may correspond to more than
one land cover type. Since the chance of a pixel to have
representation from more than two land cover types is not high, we
restrict the cardinality of the focal sets to two. Using (8) for
eight neighboring pixels we obtain eight separate bodies of
evidence. The global BPA for the focal elements is then  computed
applying Dempster's rule repeatedly. In our previous method we
could use Eq. (6) to compute the global BPA since we dealt with
singleton focal elements only. In the present case, we have to
compute the global BPA in steps, combining two bodies of evidence
at a time and preparing an intermediate BPA which will be combined
with another BPA and so on. Once the global BPA is computed, the
class label is assigned according to highest pignistic
probability.

\subsubsection{Method 4: Aggregation using evidence theoretic k-NN rule}
This method is fashioned after the evidence theoretic k-NN method
\cite{deno}. Here nine BPAs $m^0,m^1,\cdots ,m^8$ are identified
corresponding to the possibilistic label vectors
$\bm{\a}^0,\bm{\a}^1,\cdots,\bm{\a}^8$. The BPA $m^i$ is assigned
as follows: $ \hbox{ Let } q = \underbrace{arg
max}_{k}(\a^i_k),\hbox{ then}$ \beqa
m^i(\{C_q\}) &= &\a^i_q\\
m^i(\cal{C}) &= &1 - \a^i_q\\
m^i(A) &= &0\hs*{1mm}\forall A\in P(\cal{C})\backslash\{\cal{C}, \{\hbox{$C_q$} \}\}
\eeqa

Thus there is only one focal element containing one element in
each BOE. The rest of the belief is assigned to the frame of
discernment $\cal{C}$, which can be interpreted as the amount of
ignorance. The evidences are then combined using Dempster's rule
and the class label is assigned according to highest pignistic
probability.

Like Method 1 in this method also no special emphasis is given to
the pixel under consideration, $p^0$, over the neighboring pixels.
Whereas in Method 2 and Method 3, $\bm{\a}^0$ influences all the
BPAs i.e., $\bm{\a}^0$ plays a special role. It can be seen later
in experimental results that these two approaches provide
different classification efficiencies depending on the nature of
the spatial distribution of the classes. Intuitively, if pixels
corresponding to different land cover types are scattered all over
the image, neighboring pixels may not be given the same importance
as that of the central pixel for optimal performance of the
classifier. On the other hand, if pixels of a particular land
cover type form spatial clusters, then giving equal importance to
the neighboring pixels may be desirable. However, the optimal
weight to be given to the neighboring pixels to get the best
performance should depend on the distribution of different land
cover types on the image being analyzed. To realize this we modify
the BPA assignment scheme of current method as follows: $ \hbox{
Let }q = \underbrace{argmax}_{k}(\a^i_k), \hbox{ then}$ \beqa
m^0(\{C_q\}) &= &\a^0_q, \hbox{ for }i=0\\
m^i(\{C_q\}) &= &w\a^i_q, \hbox{ otherwise }\\
m^0(\cal{C}) &= &1 - \a^0_q, \hbox{ for }i=0\\
m^i(\cal{C}) &= &1 - w\a^i_q, \hbox{ otherwise }\\
m^i(A) &= &0\hs*{1mm}\forall A\in P(\cal{C})\backslash\{\cal{C}, \{\hbox{$C_q$} \}\}
\eeqa
where $0\leq w \leq 1$ is a weight factor that controls the contribution of the
neighboring pixels in the decision making process. The optimal value of $w$ for
best classification performance depends on the image under investigation and can
be learnt during training using grid search.

\section{Experimental results and discussions}
We report the performances of the proposed classifiers for two
multispectral satellite images. We call them {\bf Satimage1} and
{\bf Satimage2}. Satimage1 is a Landsat-TM image available along
with full ground truth in the catalog of sample images of the
ERDAS software and used for testing various algorithms
\cite{kumar2}. The image covers the city of Atlanta, USA and its
surrounding. Satimage2 is also a Landsat-TM image depicting
outskirts of the city of Vienna, Austria \cite{pinz}.

\begin{figure}
\epsfxsize=0.6\hsize \ArtWork{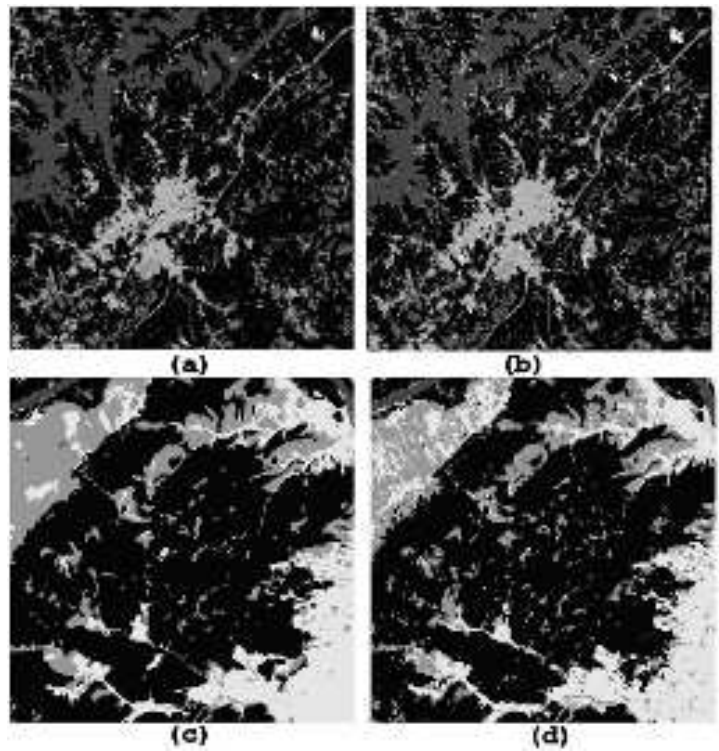} \caption{(a) The ground
truth for Satimage1. (b) The classified image for Satimage1
(Method 3, Training set 1). (c)The ground truth for Satimage2. (d)
The classified image for Satimage2 (Method 4, Training set
1).}\label{fig:gt-classif}
\end{figure}

The Sat-image1 is of size $512 \times 512$ pixels captured by
seven detectors operating in different spectral bands from
Landsat-TM3. Each of the detectors generates an image with pixel
values varying from 0 to 255. The $512 \times 512$ ground truth
data provide the actual distribution of classes of objects
captured in the image. From this data we produce the labeled data
set with each pixel represented by a 7-dimensional feature vector
and a class label. Each dimension of a feature vector comes from
one channel and the class label comes from the ground truth data.
Figure \ref{fig:gt-classif}(a) shows the ground truth for
Satimage1 where different classes are indicated with different
shades of grey.

Satimage2 also is a seven channel Landsat-TM image of size
$512\times 512$. However, due to some characteristic of the
hardware used in capturing the images the first row and the last
column of the images contain gray value 0. So we did not include
those pixels in our study and effectively worked with $511\times
511$ images. The ground truth containing four classes is used for
labeling the data. Figure \ref{fig:gt-classif}(c)  shows the
ground truth for Satimage2 where different classes are indicated
with different shades of grey.

In our study we generated 4 sets of training samples for each of
the images. For Satimage1 each training set contains 200 data
points randomly chosen from each class. For Satimage2 we include
in each training set 800 randomly chosen data points from each of
the four classes. The classifiers designed with the training data
are tested on the whole images.

The classification results of the non-contextual fuzzy rulebased
systems  are  summarized in the parts (a) and (b) of Table
\ref{tab:non-contextual} for Satimage1 and Satimage2 respectively.
Parts (a) and (b) of Table \ref{tab:contextual} summarize the
performances of the four classifiers using different methods of
aggregation of spatial information for Satimage1 and Satimage2
respectively. We used the same fuzzy rulebases for respective
training sets as used previously, but the decision making step is
modified to use the proposed four methods.
\begin{table}
\caption{Performances of non-contextual fuzzy rulebased
classifiers}\label{tab:non-contextual}
 {\footnotesize
\begin{center}
\begin{tabular}{|l|c|c|c|r|} \hline
 Trng.    &No. of     &$k_{w}$    &Error Rate in  &Error Rate in \\
 Set     &rules      &       &Training Data  &Whole Image \\ \hline
 \multicolumn{5}{|c|}{(a) Satimage1} \\ \hline
 1.      &30     &5.0        &12.0\%     &13.6\% \\ \hline
 2.      &25     &6.0        &14.3\%     &14.47\% \\ \hline
 3.      &25     &5.0        &12.0\%     &13.03\% \\ \hline
 4.      &27     &4.0        &12.6\%     &12.5\% \\ \hline
 \multicolumn{5}{|c|}{(b) Satimage2} \\ \hline
 1.      &14     &2.0        &16.3\%     &14.14\% \\ \hline
 2.      &14     &2.0        &16.3\%     &14.04\% \\ \hline
 3.      &12     &2.0        &17.09\%    &14.01\% \\ \hline
 4.      &11     &2.0        &17.34\%    &14.23\% \\ \hline
\end{tabular}
 \end{center}
 }
\end{table}
\begin{table}
\caption{Performances of fuzzy rulebased  contextual
classifiers}\label{tab:contextual}
 {\footnotesize
\begin{center}
\begin{tabular}{|l|c|c|c|c|} \hline
 Trng.         &\multicolumn{4}{c|}{Error Rate in Whole Image}\\ \cline{2-5}
 Set           &Method 1   &Method 2  &Method 3   &Method 4 \\ \hline
 \multicolumn{5}{|c|}{(a) Satimage1} \\ \hline
 1.              &13.32\%    &11.93\%    &11.48\%    &13.43\% \\ \hline
 2.              &14.00\%    &13.01\%   &12.62\%    &14.38\% \\ \hline
 3.              &13.66\%    &11.54\%   &11.23\%    &13.90\% \\ \hline
 4.              &12.98\%    &11.01\%   &10.45\%    &13.18\% \\ \hline
 \multicolumn{5}{|c|}{(b) Satimage2} \\ \hline
 1.           &11.55\%    &12.96\%    &12.64\%    &11.75\% \\ \hline
 2.           &11.55\%    &12.75\%    &12.48\%    &11.65\% \\ \hline
 3.           &11.24\%    &12.45\%    &12.23\%    &11.35\% \\ \hline
 4.           &11.24\%    &12.53\%    &12.28\%    &11.44\% \\ \hline
\end{tabular}
 \end{center}
 }
\end{table}

Comparison between Table \ref{tab:non-contextual}(a) and Table
\ref{tab:contextual}(a) shows that for Satimage1 Method 3 performs
the best with improvements varied between 1.12\% and 2.05\% and the
best performing classifier (for training set 4) achieves an error
rate as low as 10.45\%. This is closely followed by Method 2.
Methods 1 and 4 show marginal improvement for training sets 1 and 2
while for training sets 3 and 4 their performance degrades a little.
Comparison of Tables \ref{tab:non-contextual}(b) and
\ref{tab:contextual}(b) shows that although the classifiers using
Method 3 increase the classification accuracy by about 1.5\%, Method
1 is clearly the best performer for Satimage2 with improvements
ranging between 2.49\% and 2.99\%. Similar performances are achieved
by Method 4. Method 4 used $w=1$ for both Table
\ref{tab:contextual}(a) and Table \ref{tab:contextual}(b). Figure
\ref{fig:gt-classif}(b) shows the classified image corresponding to
training set 1 and Method 3 for Satimage1. Figure
\ref{fig:gt-classif}(d) displays classified image (for training set
1, Method 4) for Satimage2.

\begin{table}
  \caption{Class-wise average classification performance for the four proposed methods 1-4}\label{tab:classwise}
  {\small
  \bc
  \begin{tabular}{|l|c|c|c|c|}
    \hline
    Landcover & \multicolumn{4}{c|}{Average classification rate (\%)}\\ \cline{2-5}
    types (Frequency)           & 1  & 2  & 3 & 4 \\
    \hline
    \multicolumn{5}{|c|}{(a) Satimage1} \\ \hline
     Forest (176987)           & 93.25     & 90.80     & 91.33     & 94.47 \\ \hline
      Water (23070)             & 90.95     & 90.42     & 90.64     & 92.29 \\ \hline
     Agriculture (26986)       & 68.49     & 83.60     & 83.12     & 60.12 \\ \hline
     Bare ground (740)         & 58.15     & 65.46     & 65.92     & 55.10 \\ \hline
     Grass (12518)             & 53.19     & 73.75     & 75.65     & 55.59 \\ \hline
     Urban area (11636)       & 80.82     & 80.83     & 79.76     & 75.32 \\ \hline
     Shadow (3197)             & 48.97     & 96.09     & 95.68     & 40.42 \\ \hline
     Clouds (358)              & 99.78     & 99.78     & 99.78     & 99.79 \\
    \hline
    \multicolumn{5}{|c|}{(b) Satimage2} \\ \hline
    Built-up land (48502)     & 85.83 & 84.09 & 84.38 & 85.55 \\ \hline
    Forest (161522)           & 95.46 & 94.00 & 94.22 & 95.68 \\ \hline
    Water (2643)              & 96.02 & 93.77 & 94.32 & 96.13 \\ \hline
    Agriculture (48554)       & 68.20 & 68.03 & 67.79 & 66.92 \\
                \hline
  \end{tabular}
  \ec
  }
\end{table}

Our experimental results demonstrate that for Satimage1 Methods 3
and 2 work well while the other methods work better for Satimage2.
These differences of performances can be explained if we look into
the way the neighborhood information is aggregated in each method
and the nature of the spatial distribution of the classes in the
images. In Method 1 the fuzzy label vectors of the central pixel
(the pixel of interest) and its eight neighboring pixels are
treated in same way for aggregation. The same is true for Method 4
though evidence theoretic approach is used for information
aggregation. In Methods 2 and 3 eight BPAs are defined, each of
them is assigned using the possibilistic label vector of the
central pixel and that of one of the eight neighboring pixels.
Thus all the BPAs are heavily influenced by the central pixel, and
consequently, so is the final decision. Hence, it is expected that
for the images dominated by large stretches of homogeneous areas
(i.e., area covered by single land cover type) can be classified
better by Methods 1 and 4. Also for Method 4 there may exist an
optimal weight (different from 1) with improved performance. We
shall see later that this is indeed the case. Comparing Figure
\ref{fig:gt-classif}(c) (ground truth for Satimage2) with Figure
\ref{fig:gt-classif}(a) (ground truth for Satimage1) reveals that
Satimage1 mostly consists of small (often very small) patches of
land cover types, while Satimage2 has large patches of land cover
types. So for Satimage1 neighborhood information needs to be used
judiciously to get an improved classifier. This is what achieved
by Methods 2 and 3.

To have a closer look at class-wise performances of the proposed
methods we have presented in Table \ref{tab:classwise} the
class-wise classification performances. It can be seen from the
table that for Satimage2 all four methods perform comparably for
each of the four classes, with a slender edge in favor of Methods 1
and 4. However, for Satimage1, which contains more complicated
spatial distribution of the classes, there is significant
class-specific variation of performance among different methods. For
the classes Forest, Water, Urban area and Clouds all methods perform
nearly equally (the variation is within 5\% approximately) well,
however for other classes the performances differ significantly. For
Agriculture, the second largest class, Methods 2 and 3 have accuracy
of over 83\%, while Methods 1 and 4 are 68.49\% and 60.12\% accurate
respectively. For the class Bare ground, Methods 2 and 3 outperform
the others comfortably. For the class Shadow, there is a huge
performance gap between the Methods 2-3 ($\approx 96\%$) and the
Methods 1 and 4 (49\% and 40\%). However, since the frequency of the
Shadow class is very small, this variation does not affect the
overall accuracy significantly.

It can be observed from the above tables that in a few cases
classifiers with fewer number of rules perform better than those
with larger number of rules. This is due to the fact that each
classifier is trained with different randomly generated training
sets and there is some randomness involved in the SOFM based
prototype generation scheme. These result in different rulebases,
where one with more number of rules may have a few rules in partial
conflict.

The experiments are carried out using a desktop computer with P4
processor (3.0 GHz) and 256 MB memory. Apart from the image sizes,
the computation time depends largely on the number of rules and
number of classes. For an image the computation time can be divided
into two parts, computation of the fuzzy label vectors using
rulebase and decision making by applying the evidence-theoretic
methods. The first part is the same across all 4 methods and
requires approximately 50 seconds for Satimage1 (25-30 rules and 8
classes) and 25 seconds for Satimage2 (11-14 rules and 4 classes).
The decision making using Methods 1-4 require 2, 10, 50 and 5
seconds respectively for Satimage1 and 1, 5, 10, and 2 seconds
respectively for Satimage2. As expected, Method 1 is the fastest one
because it involves simple averaging of the label vectors. On the
other hand, Method 3 requires dealing with both singleton and
doubleton sets of classes that increase the computation time almost
quadratically with number of classes.

\begin{figure}
\epsfxsize=0.7\hsize \ArtWork{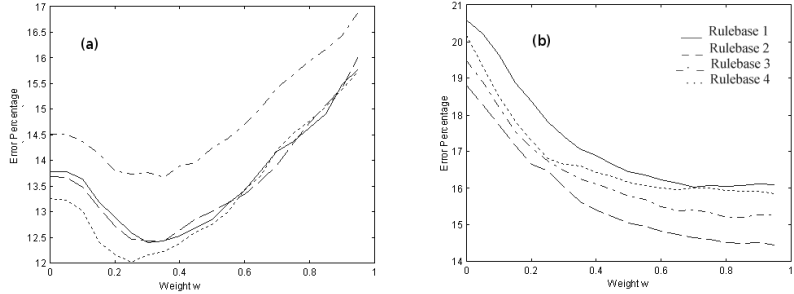} \caption{The result of
grid search for optimal value of $w$ using a $100\times 100$
sub-images of (a) Satimage1 and (b) Satimage2. The numbers beside
the plots correspond to the rulebase no.}\label{fig-gs}
\end{figure}

To investigate the possibility of optimization of Method 4, now we
use the modified Equations (12)-(16) that incorporate a weight factor
$w$ for controlling the contribution of the information from the
neighborhood in the decision making process. The value $w=1.0$ makes
the method same as the original Method 4. To find an optimal value
of $w$ we use a $100 \times 100$ sub-image from each image and find
an optimal $w$ based on these blocks. We use grid search to find the
optimal $w$. Note that, the rulebases are the same as used earlier.
For example, for Satimage1, for each of the four rule sets we find
the optimal $w$ using the classification error on the selected block
of image. Figure \ref{fig-gs}(a) depicts the variation of
classification error as a function of $w$ for Satimage1. It is
interesting to observe that for Satimage1 for all four rule bases
the best performances are achieved around $w=0.35$. So we should use
the modified Method 4 with $w=0.35$ for each of the four rulebases
and Table \ref{tab:mod-4} displays the classification errors for the
whole image. Comparing Table \ref{tab:mod-4} with column 5 of Table
\ref{tab:contextual}(b), we find a consistent improvement with
$w=0.35$ in all four cases. The improvement varied between 2.14\%
and 2.75\%. We also tried to find an optimal $w$ for each of the
four rulebases for Satimage2. Figure \ref{fig-gs}(b) displays the
classification error as a function of $w$. In this case, for all
four rulebases we find an optimal value of $w=1.0$. This again
confirms the fact that when different land cover types form
spatially compact clusters, neighbor and central pixels play equally
important roles in decision making.

\begin{table}
\caption{Classification performances of classifiers using modified
method 4 with $w=0.35$ on Satimage1}\label{tab:mod-4}
\begin{center}
\footnotesize{
\begin{tabular}{|l|c|c|c|c|} \hline
Training Set    &1      &2      &3      &4 \\ \hline
Error rate  &11.10\%    &12.24\%    &11.31\%    &10.33\% \\ \hline
\end{tabular}
}
 \end{center}
\end{table}

\subsection{Comparison with an MRF model-based classifier}
Recently contextual methods based on Markov random field (MRF)
models have become popular for classification of multispectral
\cite{sarkar} as well as multisource \cite{solberg2}, \cite{tso}
satellite images. Typically a Bayesian framework is used to model
the posterior probability. To estimate the model parameters an
energy function is minimized using an optimization technique like
simulated annealing, genetic algorithm etc. These methods usually
utilize the contextual information in the training stage. Also
their accuracy depends on the correctness of assumption about the
class-conditional density functions. Here we compare the proposed
methods with the MRF model-based approach introduced by Sarkar et
al. \cite{sarkar}. This approach is based on capturing intrinsic
characters of tonal and textural regions of a multispectral image.
Using an initially oversegmented image and the multispectral image
it defines an MRF over the region adjacency graph (RAG) of the
initial segmented image. Because of the energy minimization of the
associated MRF, the oversegmented regions are likely to be merged.
To compare the similarity of adjacent regions, a multivariate
statistical test is incorporated while  minimizing the energy
function associated with the underlying MRF.
 A cluster validation technique is also used
for finding optimal segments.

The implementation of the MRF-based scheme used for the experiments
here is capable of handling only up to four channels. Therefore, we used
channels 3, 4, 5, and 7 for Satimage1 and channels 2, 3, 5 and 7 for
Satimage2. The choice is based on visual inspection of the channel
images as well as some experiments with various combinations of
channels with a view to finding a good combination of features
suitable for the MRF-based method. Using the same training-test data sets
as used by the previous experiments, for Satimage1 the
MRF-based method fails to discriminate some of the classes, for example,
a significant part of the water becomes forest. So we used a  more detailed 10-class ground
truth data. Unfortunately this detailed ground truth is incomplete, i.e., does
not cover the entire image. Therefore, we used a training set with
1463 data points and a test set with 1480 data points. For Satimage2
the training set consisted of 3200 points as earlier. The
classification performance of the MRF-based method, non-contextual
fuzzy rulebased method and proposed contextual decision making
schemes on the test sets is summarized in Table
\ref{comp-MRF}. All experiments are conducted with the same
training and test data sets for respective images as described
in this section.

\begin{table}
  \caption{Comparison of classification performance (percentage errors)
  of MRF-based method with the proposed
  methods}\label{comp-MRF}
  \centering
  \footnotesize{
  \begin{tabular}{|c|c|c|c|c|c|c|}
    \hline
    Image & MRF & Fuzzy &\multicolumn{4}{c|}{Proposed contextual methods}
    \\\cline{4-7}
        & method    & rulebase & 1    & 2    & 3    & 4  \\    \hline
    Satimage 1     & 18.79     & 9.86           & 5.96 & 8.58 & 8.18 & 5.81 \\\hline
    Satimage 2     & 17.81     & 16.41          & 12.55 & 14.68 & 14.47 & 12.45
    \\
    \hline
  \end{tabular}
  }
\end{table}

Table \ref{comp-MRF} shows that  for Satimage1 the error rate of
the ordinary (noncontextual) fuzzy rulebased classifier is almost
half of that by the MRF-based method. All of the four proposed
contextual decision making schemes reduce the error rate further
with Method 4 achieving the lowest error rate of 5.81\%. In case
of Satimage2 also the error rate for the noncontextual fuzzy
rulebased classifier is about 2.5\% less than the MRF based
method. The proposed contextual schemes again produce further
improved classification performance. Here again Method 4 exhibits
the lowest error rate of 12.45\% closely followed by Method 1
(12.55\%). Thus it can be observed that for both the images the
ordinary fuzzy rulebased classifiers as well as the four contextual
classifiers perform consistently better than the MRF-model based
method tested here.

\section{Conclusion}
We proposed a comprehensive scheme for designing contextual
classifiers for land cover classification in multispectral satellite
images. This is a multi-stage scheme. First an SOFM-based dynamic
prototype generation algorithm is used to generate an adequate
number of prototypes. Then the prototypes are converted into fuzzy
rules and they are further fine-tuned to design an efficient fuzzy
rulebased classifier. However, this classifier is a non-contextual
one.  Since the landcover classes usually appear in spatial
clusters, a classification scheme using contextual information can
be more efficient than non-contextual one. Hence we develop four
decision making schemes which use information from spatial
neighborhood of the pixels. For assigning a class label to a pixel
the schemes use the information generated by the fuzzy rulebase in the
form of possibilistic label vectors for the pixel under consideration
as well as its
eight spatial neighbors. Three of the proposed schemes use evidence
theoretic framework and the remaining one is based on the fuzzy k-NN
rule.

The suitability of a particular scheme depends to some extent on the
nature of the image to be classified. The performance of the methods
on the training / validation  data can be used to decide on the best
classifier for a given situation. One of the methods has a parameter
that can be tuned based on the training/validation data to get a classifier
with improved performance.

\section*{acknowledgement}
The authors gratefully acknowledge the help provided by Prof. A.
Sarker to generate some results using their MRF based method. We
thank Dr. Axel Pinz to provide us with the Satimage2 data for the
experiments. We also thank the referees for their valuable
suggestions.

\ed
\begin{thebibliography}{99}
\bibitem{foody}G. M. Foody, ``Approaches for the production and evaluation of fuzzy
land cover classifications from remotely-sensed data", {\em Int. J. of Remote Sensing},
vol. 17, no. 7, pp. 1317-1340, 1996.

\bibitem{paola}J. D. Paola and R. A. Schowengerdt, ``A detailed comparison
of backpropagation neural network and maximum likelihood classifiers for
urban land use classification", {\em IEEE Trans. on Geosci. Remote Sensing},
vol. 33, pp. 981-996, July, 1995.

\bibitem{pinz}H. Bischof, W. Schneider and A. J. Pinz, ``Multispectral Classification of
Landsat-Images Using Neural Networks", {\em IEEE Trans. on Geosci.
Remote Sensing}, vol. 30, no. 3, pp. 482-490, 1992.

\bibitem{ji}C. Y. Ji, ``Land-use classification of remotely sensed datta using Kohonen
self-organizing feature map neural networks", {\em Int. J. Remote
Sensing}, vol. 18, no. 12 pp. 1451-1460, 2000.

\bibitem{foody2}G. M. Foody, ``Hard and soft classifications by a neural network with
a nonnexhaustively defined set of classes", {\em Int. J. of Remote
Sensing}, vol. 23, no. 18, pp. 3853-3864, 2002.

\bibitem{shafer}G. Shafer, {\em A Mathematical Theory  of Evidence}, Princeton University
Press, Princeton, 1976.

\bibitem{bez2}J. C. Bezdek, J. Keller, R. Krishnapuram and N. R. Pal, {\em Fuzzy Models
and Algorithms for Pattern Recognition and Image Processing} Kluwer, Massachusetts, 1999.

\bibitem{ludmila}L. Kuncheva,{\em Fuzzy Classifier Design},Physica-Verlag,2000.

\bibitem{kumar2}A. S. Kumar, S. Chowdhury and K. L. Majumder, ``Combination of neural and
statistical approaches for Classifying space-borne multispectral data,"
{\it Proc. of ICAPRDT99}, pp. 87-91, 1999.

\bibitem{bard}A. B\'{a}rdossy and L. Samaniego, ``Fuzzy rule-based classification
of remotely sensed imagery", {\em IEEE Trans. Geosci. Remote
Sensing}, vol. 40, no.2, pp. 362-374, 2002.

\bibitem{kulkarni}A. Kulkarni and S. McCaslin, ``Knowledge
discovery from multispectral satellite images", {\em IEEE Geosci.
Remote Sensing Letters}, vol. 1, no.4, pp. 246-250, 2004.

\bibitem{deno}T. Den\oe ux, ``A k-nearest neighbor classification rule based on
Dempster-Shafer theory", {\em IEEE Trans. on Syst. Man and
Cybern}, vol. 25, no. 5, pp. 804-813, 1995.

\bibitem{zouhal}L. M. Zouhal and T. Den\oe ux, ``An evidence theoretic k-NN rule with
parameter optimization", {\em IEEE Trans. on Syst. Man and Cybern:
Part C}, vol. 28, no. 2 pp. 263-271, 1998.

\bibitem{lee}T. Lee, J. A. Richards and P. H. Swain, ``Probabilistic and
evidential approaches for multispectral data analysis",{\em IEEE
Trans. Geosci. Remote Sensing}, vol. GE-25, pp. 283-293, 1987.

\bibitem{srini}A. Srinivasan and J. A. Rechards, ``Knowledge-based techniques for
multi-source classification", {\em Int. J. Remote sensing}, vol.
11, pp. 501-525, 1990.

\bibitem{peddle}D. R. Peddle, ``An empirical comparison of evidential reasoning,
linear discriminant analysis and maximum likelihood algorithms for
land cover classification", {\em Canadian J. Remote Sensing}, vol.
19, pp. 31-44, 1993.

\bibitem{kim}H. Kim and P. H. Swain, ``Evidential Reasoning Approach to
Multisource-Data Classification in Remote sensing", {\em IEEE
Trans. Syst. Man and Cybern}, vol. 25, no. 8, pp. 1257-1265, 1995.

\bibitem{jouan}A. Jouan and Y. Allard, ``Land use mapping with
evidential fusion of features extracted from polarimetric
synthetic aperture radar an hyperspectral imagery", {\em
Information Fusion}, vol 5, pp. 251-267, 2004.

\bibitem{haralick}R. M. Haralick, ``Decision making in context",
{\em IEEE Trans. Pattern Anal. Machine Intell.}, vol. PAMI-5, pp.
417-428, 1983.

\bibitem{sarkar}A. Sarkar, M. K. Biswas, B. Kartikeyan, V. Kumar,
K. L. Majumder and D. K. Pal, ``A MRF model-based segmentation
approach to classification for multispectral imagery", {\em IEEE
Trans. Geosci. Remote Sensing}, vol. 40, no. 5, pp. 1102-1113,
2002.

\bibitem{solberg2}A. H. S. Solberg, T. Taxt and A. K. Jain, ``A
markov random field model for clasification of multisouirce
satellite imagery", {\em IEEE Trans. Geosci. Remote Sensing}, vol.
34, no. 1, pp. 100-113, 1996.

\bibitem{tso}B. C. K. Tso and P. M. Mather, ``Classification of
multisource remote sensising imagery using a gentic algorithm and
markov random fields", {\em IEEE Trans. Geosci. Remote Sensing},
vol. 37, no. 3, pp. 1255-1260, 1999.

\bibitem{k1}T. Kohonen, ``The self-organizing map,"
{\em Proc. IEEE}, vol. 78, no. 9, pp. 1464-1480, 1990.

\bibitem{laha2}A. Laha and N. R. Pal ``Some novel classifiers designed using prototypes
extracted by a new scheme based on Self-Organizing Feature
Map",{\em IEEE Trans. Syst. Man and Cybern: B}, vol 31, no. 6, pp.
881-890, 2001.

\bibitem{dyck}H. Dyckhoff and W. Pedrycz, ``Generalized means as models of compensative
connectives, {\em Fuzzy Sets and Systems}, vol. 14, pp. 143-154, 1984.

\bibitem{chinmoy}N. R. Pal and C. Bose, `` Context sensitive inferencing and ``reinforcement
type" tuning algorithms for fuzzy logic systems", {\em Int. J. of
Knowledge-Based Intelligent Engineering Systems}, vol. 3, no. 4,
pp. 230-239, 1999.

\bibitem{keller}J. M. Keller, M. R. Gray and J. A. Givens, ``A fuzzy k-nearest neighbor
algorithm", {\em IEEE Trans. Syst. Man Cybern}, vol. SMC-15, no. 4, pp. 580-585, 1985.

\bibitem{smets}P. Smets and R. Kennes, ``The transferable belief model", {\em Artificial
Intelligence}, vol. 66, pp. 191-234, 1994.

\bibitem{chiu}  S. L. Chiu, "Fuzzy model identification based on cluster estimation", {\em J. Int. and Fuzzy Sys.}
, Vol. 2, pp. 267-278, 1994.

\end{thebibliography}
